\documentclass[letterpaper,10pt,conference]{ieeeconf}
\IEEEoverridecommandlockouts
\usepackage{amsmath,amssymb,booktabs,graphicx,cite,url}
\graphicspath{{figures/}}
\newcommand{\E}{\mathbb{E}}
\newcommand{\sg}{\operatorname{sg}}
\newcommand{\clip}{\operatorname{clip}}
\newcommand{\ours}{ACT--Flow}
\title{Learning to Absorb Corrections: Critic-Weighted Residual Flow\\for Vision-Based Insertion}
\author{Yiqing Yuan$^{1}$, Zhi Li$^{1,2}$, Jinrong Yang$^{1,2,\dagger}$, Mingshan Sun$^{2}$,\\
Mei Yu Zhi$^{2}$, Yong Zhao$^{2}$, Liangliang Ren$^{2}$, and Hui Cheng$^{1,*}$%
\thanks{$^{1}$Sun Yat-sen University. $^{2}$CVTE.}%
\thanks{$^{\dagger}$Project leader: Jinrong Yang. $^{*}$Corresponding author: Hui Cheng.}}
\begin{document}
\maketitle
\thispagestyle{empty}
\pagestyle{empty}

\begin{abstract}
Vision-based insertion requires correcting alignment errors while retaining useful behavior learned from demonstrations. We present \ours, an online learning framework that couples an adaptive Action Chunking Transformer (ACT) with a conditional residual flow policy. Following anchor-guided warmup, ACT absorbs successful executed actions through behavioral cloning, while residual targets are recomputed relative to the updated prior. A chunk-level critic learns from successful and failed interactions and evaluates final composed actions. Its estimates weight generated residual candidates for flow matching, complemented by supervision from successful real executions. Candidate scoring is confined to training; deployment uses RGB observations and samples one residual per action chunk. We evaluate the framework on power-plug, USB, and key insertion, together with cable routing and needle threading. Results show a five-task mean success rate of 66\% at 20,000 post-initialization interaction steps, exceeding the strongest baseline by 18 percentage points at this checkpoint. Evaluations at matched checkpoints with the residual disabled show ACT's mean success increasing from 16\% to 76\%. Component ablations further support the complementary roles of behavior absorption, value weighting, and successful-action supervision.
\end{abstract}

\section{Introduction}
Vision-based insertion requires a robot to correct alignment errors during execution. Inserting a power plug, USB connector, or key can fail when small position or orientation errors persist, or when the target moves after the initial approach. Demonstration-trained policies, including Action Chunking Transformer (ACT)~\cite{act} and Diffusion Policy~\cite{diffusionpolicy}, provide effective action priors. However, limited demonstrations cannot cover every state encountered during deployment, and deviations from the training distribution can compound~\cite{dagger}. This motivates a policy that can both correct execution errors and incorporate successful corrections into its learned behavior.

\begin{figure}[t]
  \centering
  \includegraphics[width=\columnwidth]{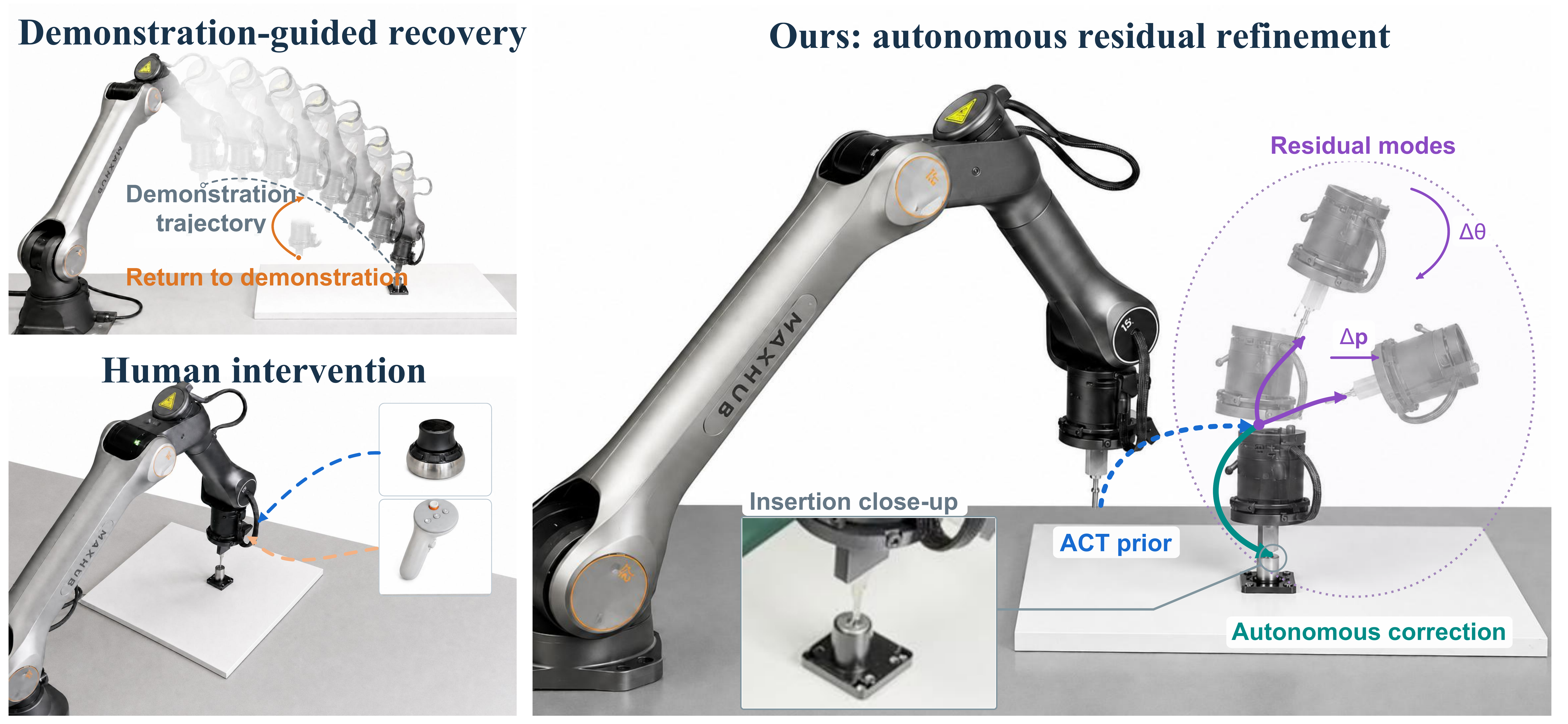}
  \caption{Learning and absorbing corrections. Demonstration-guided recovery and human intervention provide corrective guidance. Our framework initializes ACT and a residual flow policy through anchor-guided demonstration and correction warmup. During subsequent autonomous interaction, a critic trained on successful and failed experience guides residual learning, while successful executed actions update ACT to absorb corrected behavior.}
  \label{fig:intro}
\end{figure}

Existing methods obtain corrective guidance from demonstrations or human intervention. DOME~\cite{dome}, MILES~\cite{miles}, and AutoSERL~\cite{autoserl} use demonstrations to guide visual alignment, recovery, or automatic intervention. HIL-SERL~\cite{hilserl} incorporates human corrections into reinforcement learning (RL), while CR-DAgger~\cite{crdagger} learns residual policies from compliant corrections. Demonstration-guidance mechanisms depend on available reference behavior, while human corrections require operator effort. Beyond obtaining guidance, a further question is how the robot should represent corrections and retain the useful behavior they produce.

Residual adaptation provides a natural representation: a base policy supplies nominal behavior and an additional policy learns corrections~\cite{residualrl,crdagger,ftc,hilresrl}. FlowDAgger~\cite{flowdagger} similarly steers a frozen generative policy through its latent variables. A frozen prior provides a stable basis for adaptation, but successful corrections remain in the additional module rather than becoming part of the prior itself. The central question is: \emph{how can a robot learn effective insertion corrections from interaction and absorb successful corrected behavior into an improving action prior?}

Joint prior--residual learning raises two issues. Updating the prior changes the residual needed to reproduce a successful action, making historical labels inconsistent. Critic estimates rank corrections but cannot replace observed successful actions as supervision. Learning must combine these signals while tracking the evolving prior. Retaining successful executed actions provides a stable reference: their decomposition can change as the prior improves, while their role as observed targets is preserved.

We present \ours, an online framework for vision-based insertion that coordinates an adaptive ACT prior with a conditional residual flow policy. Following anchor-guided demonstration and correction warmup, successful executed actions update ACT through behavioral cloning (BC), while residual targets are recomputed against the updated ACT to preserve consistent supervision. To combine value guidance with observed successful behavior, residual learning couples critic-weighted flow matching (FM)~\cite{flowmatching} on generated candidates with FM on successful executions. The critic learns from both outcomes and scores composed actions.

We evaluate three insertion tasks---power-plug, USB, and key insertion---plus cable routing and needle threading. Comparisons and ablations examine adaptation, behavior absorption, and residual supervision. Our contributions are:
\begin{itemize}
  \setlength{\itemsep}{2pt}
  \setlength{\parskip}{0pt}
  \setlength{\parsep}{0pt}
  \item An online learning framework for vision-based insertion connecting an ACT prior and residual flow policy through anchor-guided initialization, online correction, and successful-experience feedback.
  \item A behavior-absorption mechanism that updates ACT from successful actions and relabels residual targets under the current prior, keeping correction supervision consistent as the base policy changes.
  \item Critic-weighted residual flow learning that combines value-weighted fitting of generated corrections with FM supervision from successful executed actions.
  \item Comparisons on three real-world insertion tasks and two additional manipulation tasks, supported by component ablations and diagnostics of base-policy improvement, residual contribution, and critic ranking quality.
\end{itemize}

\section{Related Work}
\subsection{Demonstrations and Corrective Interaction}
DAgger~\cite{dagger} aggregates corrective supervision, while CR-DAgger~\cite{crdagger} collects compliant corrections. MILES~\cite{miles} augments a reference demonstration through self-supervised recovery, and HD-Space~\cite{hdspace} organizes collection within overlapping atomic state--action spaces. Simulation can also supply demonstrations: Li et al.~\cite{conveyorgem} use appearance-noise annealing to shape optimization for sim-to-real conveyor manipulation. SERL~\cite{serl} supports practical real-world RL; HIL-SERL~\cite{hilserl} incorporates human corrections, while AWAC~\cite{awac} and RLPD~\cite{rlpd} exploit prior data for online learning. AutoSERL~\cite{autoserl} automates demonstration-guided intervention and recovery, and Focus-Then-Contact (FTC)~\cite{ftc} combines residual RL with affordance guidance and timed human intervention.

Visual task conditioning offers a complementary route to robust imitation. Vision-instructed pre-training (VIP)~\cite{vipvision} uses future images and sparse point flows to specify targets; SAM2Grasp~\cite{sam2grasp} combines object prompts with temporal tracking to disambiguate grasp trajectories. These approaches clarify task targets, while our focus is learning and absorbing corrections during interaction.

In our setting, corrective experience must support both prior improvement and residual learning. We use HD-Space's collection principle for warmup, successful actions to update ACT, and both outcomes to train the residual-learning critic.

\subsection{Residual Adaptation and Prior Updating}
Residual RL~\cite{residualrl} adds learned corrections to a nominal controller. Policy Decorator~\cite{policydecorator} refines a frozen imitation policy and evaluates composed actions. CR-DAgger~\cite{crdagger} learns residual motions and forces from compliant supervision; FTC~\cite{ftc} and HiL-ResRL~\cite{hilresrl} combine frozen priors with human-guided residual RL. FlowDAgger~\cite{flowdagger} steers a frozen generative policy through latent-space corrections. In contrast, RL-100 directly fine-tunes an imitation-pretrained diffusion policy through offline and online RL~\cite{rl100}.

A frozen prior provides a stable residual reference but does not absorb corrections. Updating the prior permits behavior absorption, yet changes the residual required to reproduce a previously successful action. We retain executed actions as stable targets, update ACT from successful experience, and recompute residual targets against the current ACT.

\subsection{Value-Guided Generative Policy Learning}
Advantage-weighted regression~\cite{awr} uses value-derived weights for supervised policy fitting. FM~\cite{flowmatching} provides a generative regression objective; reward-weighted fine-tuning~\cite{weightedfm} favors high-reward samples while regularizing deviation from a reference model. Flow Q-Learning~\cite{fql} trains a separate one-step actor with value optimization and distillation from a behavior-cloned flow policy. PACT~\cite{pact} calibrates credit assignment using intervention-derived preferences, while Max-Q~\cite{maxq} uses chunk-value regression to select between policy and replay actions for imitation.

Under an evolving prior, critic rankings provide value guidance but cannot replace successful-action targets. We combine critic-weighted FM on generated candidates with successful-execution FM, both referenced to the current ACT.

\section{Problem Formulation}
\label{sec:problem}
We consider RGB-based insertion: a robot aligns an object with a target and completes insertion. Let $t$ index control steps. At a decision boundary, policy $\pi$ receives observation $o_t$ and outputs a normalized action chunk $A_t\in[-1,1]^{H\times d}$ of $H$ commands, each with $d$ components. After $k\leq H$ executed steps, the robot receives $o_{t+k}$; early termination gives $k<H$.

Starting from demonstrations $\mathcal D_0$, the robot learns within a finite interaction budget. Let $p_t\in\mathbb R$ estimate task progress, with larger values indicating greater completion. Feedback combines progress differences and terminal success:
\begin{equation}
 r_t^{\mathrm{prog}}=p_{t+1}-p_t,\qquad r_t=r_t^{\mathrm{prog}}+100\,s_t,
 \label{eq:reward}
\end{equation}
Here $r_t^{\mathrm{prog}}$ is the figure's process reward; $s_t=1$ only on successful termination and is zero otherwise. Progress increases yield positive feedback and decreases yield negative feedback; success adds 100 once. The estimator is defined in Sec.~\ref{sec:progress_critic}.

For a fixed progress estimator, we maximize expected discounted return within the interaction budget:
\begin{equation}
 \max_\pi J(\pi)
 =\E_{\xi\sim\pi}\!\left[\sum_{t=0}^{T_{\mathrm{ep}}-1}\gamma^t r_t\right],
 \label{eq:objective}
\end{equation}
Here $\xi$ is a trajectory under $\pi$, $T_{\mathrm{ep}}$ its length in control steps, and $\gamma\in(0,1)$ the discount factor. Evaluation uses task success rate.

\begin{figure*}[t]
  \centering
  \includegraphics[width=\textwidth]{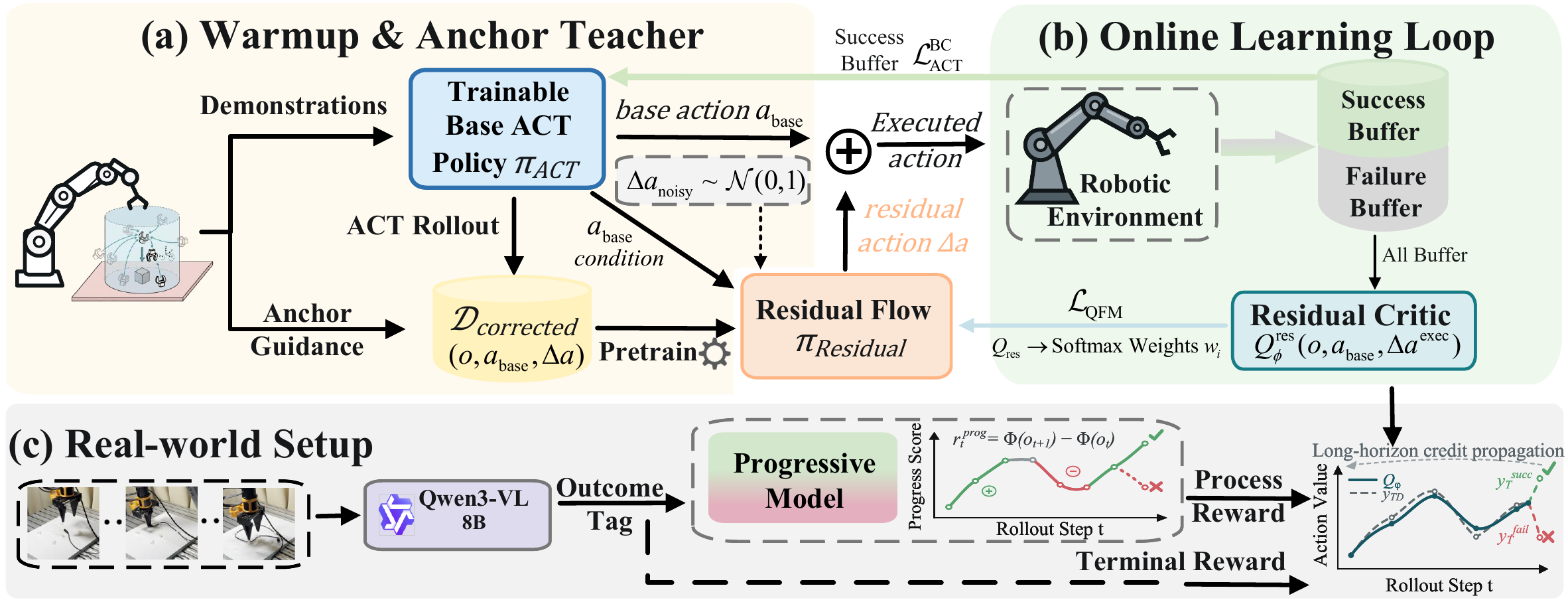}
  \caption{\ours\ pipeline. (a) Anchor demonstrations initialize ACT; corrected executions warm up the residual flow. (b) Successful actions update ACT and supply residual targets recomputed against the current prior. A critic trained on both outcomes weights composed candidates for critic-weighted FM (QFM), complemented by FM on successful executions. (c) The ReWiND-based Progressive Model~\cite{rewind} is fine-tuned during warmup with Qwen3-VL-8B terminal-success supervision and frozen online. Progress differences and terminal success provide value-learning rewards.}
  \label{fig:pipeline}
\end{figure*}

\section{Method}
\label{sec:method}
Figure~\ref{fig:pipeline} summarizes \ours: an ACT prior~\cite{act} and conditional residual flow improve together while retaining consistent supervision. Sections~\ref{sec:absorption}--\ref{sec:progress_critic} describe initialization and absorption, residual learning, and progress and value estimation, respectively.

The executed action combines the prior and residual:
\begin{equation}
 \begin{aligned}
 a_{\mathrm{base},t} &= \clip(\pi_{\mathrm{ACT},\psi}(o_t),-1,1),\\
 \Delta a_t &= \pi_{\mathrm{Residual},\theta}(o_t,\sg(a_{\mathrm{base},t}),\Delta a_{\mathrm{noisy}}),\\
 &\hspace{1em}\Delta a_{\mathrm{noisy}}\sim\mathcal N(0,I),\\
 A_t &= \clip(a_{\mathrm{base},t}+\lambda\Delta a_t,-1,1),\quad \lambda>0.
 \end{aligned}
 \label{eq:compose}
\end{equation}
Here $\psi,\theta$ parameterize the figure's $\pi_{\mathrm{ACT}},\pi_{\mathrm{Residual}}$. Base, residual, and noise are $H\times d$ chunks; noise entries are independent standard normals ($I$ is the flattened identity covariance). The operators $\clip$ and $\sg$ denote elementwise clipping and stop-gradient. The fixed scale $\lambda$ and clipping complete the figure's addition operation; $A_t$ is the legal composed chunk. Raw residuals are unrestricted. This composition is shared by warmup, collection, scoring, and value targets; zero residual recovers ACT. We omit $t$ for individual training samples. Multiple residual candidates are generated for critic-weighted flow updates, whereas each interaction step at a chunk boundary uses a single residual sample.

\subsection{Anchor Warmup and Behavior Absorption}
\label{sec:absorption}
Anchor warmup initializes the policy; online updates then alternate behavior absorption and residual relabeling.

\subsubsection{Anchor-guided warmup}
We define anchors as an ordered set of task-relative end-effector reference poses associated with successive manipulation stages. For insertion, these anchors specify an approach pose, an aligned pre-insertion pose, and a terminal insertion pose. For cable routing, they specify successive cable-placement poses along the desired route, including the entry and exit of each clip.

Following HD-Space~\cite{hdspace}, we collect demonstrations across overlapping atomic state--action spaces with varied stage-local starting configurations. The anchor teacher supplies $\mathcal D_{\mathrm{anchor}}=\mathcal D_0$ for ACT BC; its privileged geometry and target information are excluded from policy and critic inputs. We freeze the resulting parameters $\psi_0$ and execute anchor-corrected chunks at observations encountered by ACT.

Let $\mathcal D_{\mathrm{corrected}}^+$ contain successful, fully executed correction chunks $A^{\mathrm{exec}}$. With $a_{\mathrm{base}}^{(0)}=\clip(\pi_{\mathrm{ACT},\psi_0}(o),-1,1)$, where $(0)$ identifies the frozen initial prior, the teacher residual and warmup loss are
\begin{equation}
 \begin{aligned}
 \Delta a^{\mathrm{teach}}&=(A^{\mathrm{exec}}-a_{\mathrm{base}}^{(0)})/\lambda,\\
 \mathcal L_{\mathrm{warm}}&=\E_{\mathcal D_{\mathrm{corrected}}^+}
 [\ell_{\mathrm{FM}}(\theta;(o,a_{\mathrm{base}}^{(0)}),\Delta a^{\mathrm{teach}})],
 \end{aligned}
 \label{eq:warmup}
\end{equation}
The superscript $+$ selects successful data; $\ell_{\mathrm{FM}}$ is the single-target FM loss in Sec.~\ref{sec:residual_flow}. Only Flow is updated here. We then pretrain the critic on observed warmup transitions using the composed policy for bootstrapping, and evaluate before online collection. Anchor guidance ends after initialization; its demonstrations remain available for rehearsal.

\subsubsection{Online behavior absorption}
Each ACT update combines anchor rehearsal with successful online data:
\begin{equation}
 \mathcal L_{\mathrm{ACT}}^{\mathrm{BC}}=
 \alpha\mathcal L_{\mathrm{BC}}(\mathcal D_{\mathrm{anchor}})
 +(1-\alpha)\mathcal L_{\mathrm{BC}}(\mathcal D_{\mathrm{online}}^+),
 \label{eq:act}
\end{equation}
Here $\mathcal L_{\mathrm{BC}}(\mathcal D)$ is ACT's BC loss, $\alpha\in[0,1]$ sets the mixture, and $\mathcal D_{\mathrm{online}}^+$ contains successful online executions. We retain ACT's reconstruction and latent regularization~\cite{act}, masking unexecuted suffixes. Anchor rehearsal preserves demonstration supervision while the online term absorbs corrected behavior; ACT receives only BC gradients. The online targets are actions actually executed by the combined policy, allowing improvements initially supplied by the residual to become reproducible by ACT itself.

\subsubsection{Residual relabeling and replay}
After updating ACT, we recompute the successful residual target $\Delta a^+$ for each complete executed action:
\begin{equation}
 \Delta a^+=\sg\!\left(
 \frac{A^{\mathrm{exec}}-\clip(\pi_{\mathrm{ACT},\psi}(o),-1,1)}{\lambda}
 \right).
 \label{eq:relabel}
\end{equation}
Executed actions remain fixed targets as the prior changes: composing the updated base with $\lambda\Delta a^+$ reconstructs $A^{\mathrm{exec}}$. The same successful action generally requires a different residual after an ACT update. Reusing its historical residual would supervise a correction relative to an outdated base. Each batch therefore recomputes targets under the updated ACT, held fixed for both Flow objectives.

The critic view $\mathcal D_Q$ contains successful and failed warmup/online transitions with complete planned chunks and actual execution prefixes. The Flow view $\mathcal D_F$ selects fully executed successful chunks, whereas ACT also uses masked prefixes. A chunk truncated by successful termination can still train ACT through its executed prefix and train the critic through the observed transition, without providing an incomplete Flow target. Observations from $\mathcal D_Q$, including failed attempts, provide contexts for candidate generation.

\subsection{Critic-Weighted Residual Flow Learning}
\label{sec:residual_flow}
The residual policy combines QFM on model-generated candidates with successful-action FM (SFM) on real executions. QFM emphasizes promising corrections; SFM supplies observed targets independent of critic rankings.

We parameterize the residual generator with a conditional vector field $v_\theta(x,u\mid c)$, where $c=(o,\sg(a_{\mathrm{base}}))$ contains the observation and current base chunk. Chunks are flattened into $Hd$ coordinates for flow learning. Here $u$ is flow time, distinct from environment time $t$; $x_0$, $x_1$, and $x_u$ are the noise start, target residual, and interpolated residual, respectively. Following FM~\cite{flowmatching}, for a target residual $x_1$ we draw $x_0\sim\mathcal N(0,I)$ and $u\sim\mathcal U[0,1]$, and construct $x_u=(1-u)x_0+ux_1$. FM fits the vector field to velocity $x_1-x_0$:
\begin{equation}
 \begin{aligned}
 \ell_{\mathrm{FM}}(\theta;c,x_1)
 =\frac{1}{Hd}\E_{x_0,u}\big[&\|v_\theta(x_u,u\mid c)\\
 &-(x_1-x_0)\|_2^2\big].
 \end{aligned}
 \label{eq:fm}
\end{equation}
At sampling time, integrating the vector field from Gaussian noise over $u\in[0,1]$ defines the residual policy $\pi_{\mathrm{Residual},\theta}(o,a_{\mathrm{base}},\Delta a_{\mathrm{noisy}})$. The numerical integration budget $K$ controls the cost of generating a residual and is distinct from the environmental execution horizon $H$.

For value guidance, let $Q_{\mathrm{score}}$ be the frozen chunk-value estimator learned in Sec.~\ref{sec:progress_critic}. Before each flow gradient step, a no-gradient snapshot $\theta_{\mathrm{ref}}$ of the current flow produces $N$ independent candidates at the same observation, indexed by $i=1,\ldots,N$:
\begin{equation}
 \begin{aligned}
 \Delta a_i &= \sg(\pi_{\mathrm{Residual},\theta_{\mathrm{ref}}}(o,a_{\mathrm{base}},\Delta a_{\mathrm{noisy},i})),
 \\[-2pt]
 &\hspace{1em}\Delta a_{\mathrm{noisy},i}\sim\mathcal N(0,I),\\
 q_i &= \sg\!\left(Q_{\mathrm{score}}
 \big(o,\clip(a_{\mathrm{base}}+\lambda\Delta a_i,-1,1)\big)\right).
 \end{aligned}
 \label{eq:candidates}
\end{equation}
The critic scores composed, clipped actions, while raw residuals $\Delta a_i$ remain FM endpoints. During policy training, candidate generation and scoring require no additional robot executions and create no replay transitions.

We convert these scores into weights within each observation's candidate group:
\begin{equation}
 w_i=\frac{\exp((q_i-q_{\max})/\tau)}
 {\sum_{j=1}^{N}\exp((q_j-q_{\max})/\tau)},
 \label{eq:weights}
\end{equation}
Here $\tau>0$ controls weight concentration and $q_{\max}=\max_jq_j$ stabilizes computation. Normalization is within each observation's candidate group, giving higher-value corrections more regression weight~\cite{awr,weightedfm}.

The QFM term fits critic-weighted candidate residuals, while the SFM term fits residual targets derived from successful executed actions. Their joint objective is
\begin{equation}
 \begin{aligned}
 \mathcal L_{\mathrm{QFM}}&=\E_{o\sim\mathcal D_Q}
 \left[\sum_{i=1}^{N}w_i\ell_{\mathrm{FM}}(\theta;c,\Delta a_i)\right],\\
 \mathcal L_{\mathrm{SFM}}&=\E_{(o,A^{\mathrm{exec}})\sim\mathcal D_F}
 \left[\ell_{\mathrm{FM}}(\theta;c,\Delta a^+)\right],\\
 \mathcal L_{\mathrm{Flow}}&=\mathcal L_{\mathrm{QFM}}
 +\beta\mathcal L_{\mathrm{SFM}},\qquad \beta\geq0.
 \end{aligned}
 \label{eq:flow}
\end{equation}
Here $\beta$ weights SFM and $\mathcal L$ averages the single-target loss $\ell_{\mathrm{FM}}$. Equation~\eqref{eq:relabel} supplies $\Delta a^+$. QFM learns candidate corrections at successful and failed states; SFM learns from actual successful actions. The critic weights alternative corrections by predicted return, while successful executions anchor learning to actions that have completed the task. Failed attempts thus supply value evidence and candidate contexts without becoming successful-action imitation targets. Joint updates require both data views to be available.

We fit candidates separately, averaging losses rather than actions. Distinct candidates may encode different corrective motions; averaging their actions could produce an intermediate target that matches neither correction. FM noise and interpolation times are independent of candidate-generation noise. Base contexts, endpoints, scores, and weights are detached; gradients update only the vector field. Candidate snapshots refresh after each step, while ACT and $Q_{\mathrm{score}}$ remain fixed throughout the Flow phase.

\subsection{Progress Estimation and Chunk Value Learning}
\label{sec:progress_critic}
Our ReWiND-based Progressive Model~\cite{rewind} predicts $p_t=\Phi_\omega(I_t,\ell)$ from a single RGB image and task description, unlike the value-implicit VIP reward model's goal-image conditioning~\cite{vip}. Figure~\ref{fig:pipeline} abbreviates this as $\Phi(o_t)$. Language conditions reward estimation only. Qwen3-VL-8B~\cite{qwen3vl} supplies terminal success labels $s_t$ during warmup and online learning, with 97\% classification accuracy against human annotations. Following ReWiND's training and normalization procedures, we fine-tune $\Phi_\omega$ on the 30 warmup episodes and freeze it during online learning. Progress differences and terminal success define Eq.~\eqref{eq:reward}.

We use a chunk critic $Q_\phi(o_t,A_t)$ to evaluate the complete composed action from Eq.~\eqref{eq:compose}. The residual-coordinate notation in Fig.~\ref{fig:pipeline} denotes the same value through composition:
\[
\begin{aligned}
Q_\phi^{\mathrm{res}}(o,a_{\mathrm{base}},\Delta a)
&:=Q_\phi(o,A),\\
A&=\clip(a_{\mathrm{base}}+\lambda\Delta a,-1,1).
\end{aligned}
\]
Here $\mathrm{res}$ identifies residual learning; the network evaluates the composed action $A$. The figure's $\Delta a^{\mathrm{exec}}$ denotes its recorded correction. Replay stores planned chunks, executed prefixes, rewards, termination flags, and next observations.

For $k\leq H$ executed steps, the chunk return $R_t$ and discount $\Gamma_t$ are
\begin{equation}
 R_t=\sum_{j=0}^{k-1}\gamma^j r_{t+j},
 \qquad \Gamma_t=\gamma^k.
 \label{eq:chunk_return}
\end{equation}
This includes the terminal bonus exactly once at its control-step discount. When $k<H$, both the accumulated reward and $\gamma^k$ reflect the observed execution duration, so early termination remains consistent with the original chunk decision. Let $h\in\{1,2\}$ index critic heads, with online parameters $\phi_h$ and target parameters $\bar\phi_h$. Using the bootstrap mask $m_t$, we form the TD target $y_t^{\mathrm{TD}}$ and critic loss:
\begin{equation}
 \begin{aligned}
 y_t^{\mathrm{TD}} &= R_t+\Gamma_t m_t\,
 \frac{1}{2}\sum_{h=1}^{2}Q_{\bar\phi_h}(o_{t+k},A'),\\
 \mathcal L_Q &=\frac{1}{2}\sum_{h=1}^{2}\E_{\mathcal D_Q}
 \left[\big(Q_{\phi_h}(o_t,A_t)-\sg(y_t^{\mathrm{TD}})\big)^2\right].
 \end{aligned}
 \label{eq:critic}
\end{equation}
Here $m_t=0$ at true termination and $m_t=1$ when bootstrapping is permitted; time-limit truncation follows the fixed task-specific bootstrap rule. The next action $A'$ is sampled using frozen snapshots of the updated ACT and the flow before its update in the current round. Target heads track their online heads by Polyak averaging during critic training. Both the target value and the subsequent scoring model use the mean of the two target heads. The figure's $y_{\mathrm{TD}}$ is this target; $y_T^{\mathrm{succ}}$ and $y_T^{\mathrm{fail}}$ denote its terminal cases at boundary $T$, including progress rewards.

\begin{figure*}[t]
  \centering
  \includegraphics[width=\textwidth]{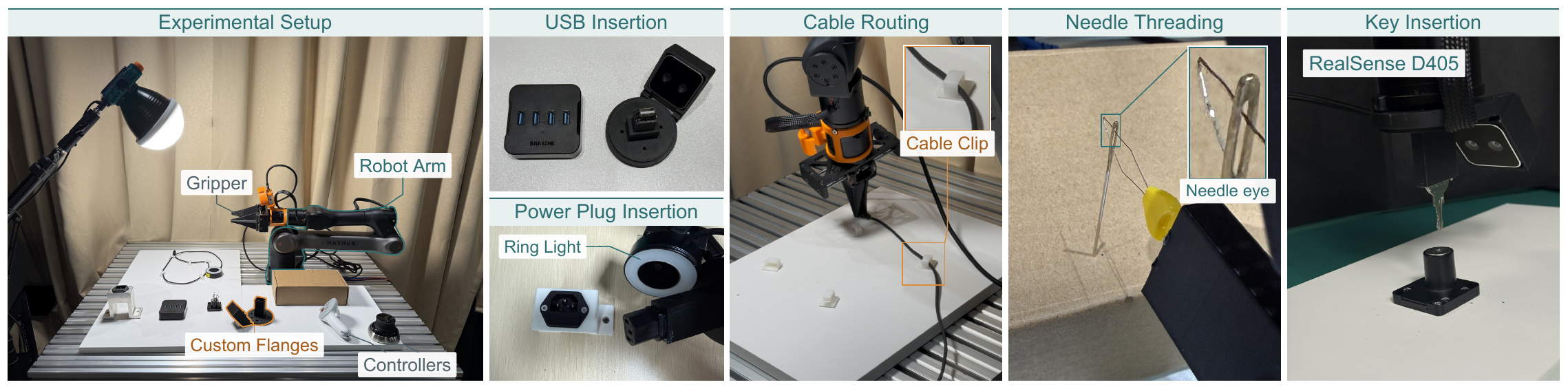}
  \caption{Real-world platform and five tasks. The setup includes a six-axis robot arm, gripper, custom task-specific flanges, controllers for baseline data collection, a work light and ring light, and a RealSense D405 close-view camera. Task panels show USB and power-plug insertion, cable routing, needle threading, and key insertion.}
  \label{fig:setup}
\end{figure*}

All valid successful and failed transitions train $\mathcal L_Q$. The critic takes only observations and planned legal chunks; $k$ enters return and bootstrap construction. Its value estimates discounted task return. After critic training, we freeze
\begin{equation}
 Q_{\mathrm{score}}(o,A)=\frac{1}{2}
 \sum_{h=1}^{2}Q_{\bar\phi_h}(o,A)
 \label{eq:scorer}
\end{equation}
for the candidate-weighted updates of Sec.~\ref{sec:residual_flow}.

Each round collects with a fixed ACT--Flow pair, updates ACT from successful executions, trains the critic using the updated ACT and pre-update Flow, then optimizes Eq.~\eqref{eq:flow} with fixed critic scores and current-base residual targets. This order aligns residual targets and critic evaluation with the current prior, while keeping the scoring function fixed during each Flow update. The updated pair collects the next round.

\section{Experiments}
\label{sec:experiments}
We evaluate \ours\ on five real-world tasks, including three insertion tasks, to examine online learning efficiency, successful-behavior absorption, and the value of critic-guided residual learning. Comparisons assess overall performance, while ablations and paired diagnostics explain the contribution of each component.

\begin{figure}[!b]
  \centering
  \includegraphics[width=\columnwidth]{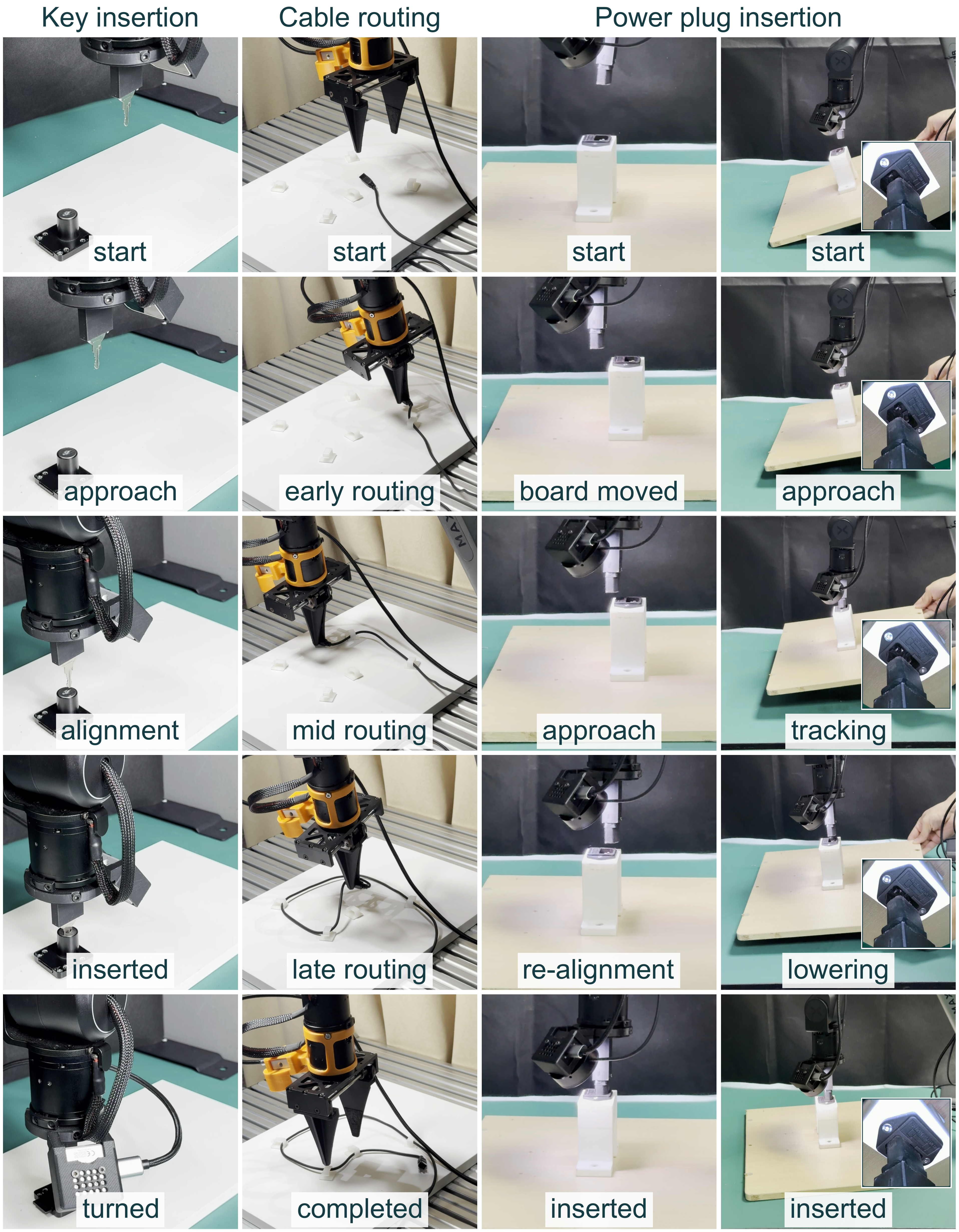}
  \caption{Real-world executions from top to bottom. Columns show key insertion, cable routing, rapid target re-localization after socket-board repositioning, and pose following as the board is lowered.}
  \label{fig:trajectories}
\end{figure}

\begin{figure*}[t]
  \centering
  \includegraphics[width=\textwidth]{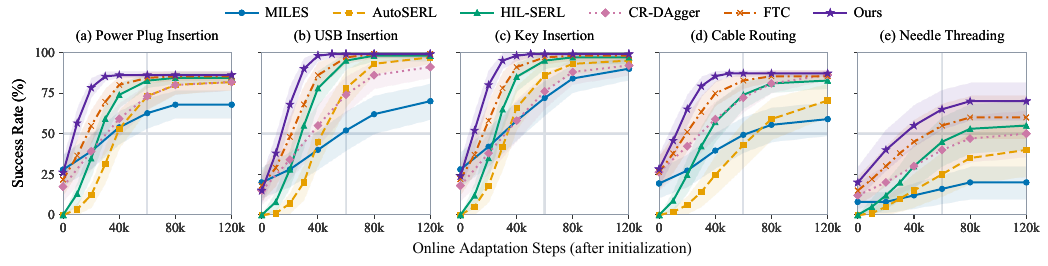}
  \caption{Main learning curves across five tasks. Online environment steps are counted after 30 initialization episodes. Curves and bands show the three-seed means and standard deviations (SDs). Lines connect discrete evaluations.}
  \label{fig:main}
\end{figure*}

\subsection{Environmental Setup}
\label{sec:experimental_setup}
\textbf{Platform and tasks:} Figure~\ref{fig:setup} shows the six-axis arm, gripper, task-specific flanges, and RealSense D405 close-view camera. Work and ring lights illuminate the table and contact region. Collection controllers serve the baselines; \ours\ uses anchor-guided warmup (Sec.~\ref{sec:absorption}).

Power-plug, USB, and key insertion require connector--socket or key--keyway alignment. Cable routing places a deformable cable through clips; needle threading passes flexible thread through an eye. Task completion defines success.

\textbf{Initialization and training:} The comparison protocol allocates 30 episodes to initialization and defines online step zero after initialization training. Our warmup collection follows HD-Space~\cite{hdspace} and the staged procedure in Sec.~\ref{sec:absorption}. Anchor poses are manually specified relative to the socket or cable fixtures before warmup and remain fixed in their respective local frames. The learning curves then cover up to $120$k real online environment steps. We use chunk horizon $H=4$, flow integration budget $K=10$, $N=8$ training candidates, $\lambda=1$, two critic heads, $\gamma=0.99$, and target update rate $0.005$. Initial values are $\tau=1$, $\beta=1$, and equal anchor/online-success sampling for ACT. Candidate generation and scoring use model computation; only actual robot execution contributes online environment steps.

\textbf{Baselines:} We compare with five methods. MILES~\cite{miles} augments a demonstration with self-supervised recovery trajectories. AutoSERL~\cite{autoserl} constructs automatic intervention and recovery from demonstration-derived guidance. HIL-SERL~\cite{hilserl} incorporates human corrections into real-world RL. CR-DAgger~\cite{crdagger} learns compliant residual corrections from human supervision. FTC~\cite{ftc} combines affordance-guided residual RL with timed human intervention. All methods are evaluated on the same tasks. Online environment steps are counted after initialization.

\textbf{Evaluation metrics:} At each checkpoint, we evaluate 100 episodes per task per training seed, with success judged by humans. Main learning curves report means and standard deviations (SDs) of success rate (SR) over three training seeds. Lines connect evaluated checkpoints. For Table~\ref{tab:ablation}, final SR is averaged over the last three predetermined evaluation checkpoints. Normalized area under the learning curve (NAUC) measures performance throughout adaptation. For success $p_j\in[0,1]$ at step $s_j$, it is computed as
\begin{equation}
 \mathrm{NAUC}=\frac{1}{B}\sum_{j=1}^{J}
 \frac{p_j+p_{j-1}}{2}(s_j-s_{j-1}),\quad B=s_J.
 \label{eq:nauc}
\end{equation}
Metrics are computed per task and seed before equal task averaging; the table reports a $0$--$100$ scale. Means, SDs, and differences are computed before rounding reported results to integers. NAUC uses all evaluations; absorption diagnostics use four checkpoints.

\subsection{Comparative Experiments}
\label{sec:comparisons}
\textbf{Online learning efficiency:} Figure~\ref{fig:main} shows the clearest advantage of \ours\ during early adaptation. At $20$k online steps, it reaches $78\%$, $68\%$, $80\%$, $65\%$, and $40\%$ on power-plug insertion, USB insertion, key insertion, cable routing, and needle threading, respectively. It has the highest mean on each task at this checkpoint. Its five-task mean is $66\%$, compared with $48\%$ for FTC, the strongest baseline at this stage, giving an $18$-percentage-point difference. The corresponding means for MILES, AutoSERL, HIL-SERL, and CR-DAgger are $29\%$, $10\%$, $27\%$, and $35\%$. Gains across rigid and deformable tasks support combining demonstrated priors with online correction.

\textbf{Performance at larger budgets:} The average gap to FTC narrows from $18$ points at $20$k to $9$ at $40$k and $4$ at $60$k. By $60$k steps, \ours\ reaches a mean SR of $87\%$, close to its $88\%$ endpoint at $120$k. The final task rates are $86\%$, $99\%$, $99\%$, $87\%$, and $70\%$, respectively. Several baselines approach similar late performance on insertion tasks, while needle threading retains a larger separation: its $70\%$ mean exceeds FTC's $60\%$ by ten points. Threading also has wider variability ($12$ and $13$ points SD, respectively). Together, these results demonstrate faster early adaptation, near-final mean performance by $60$k steps, and a sustained advantage on needle threading.

\textbf{Representative executions:} Figure~\ref{fig:trajectories} shows key insertion and turning, and cable placement through clips.

The power-plug sequences distinguish two corrections. The third column shows \emph{rapid target re-localization}: after socket-board displacement, the robot redirects its approach and re-aligns for insertion. The rightmost column shows \emph{pose following}: as the initially inclined board is lowered, the end effector follows the socket's changing position and orientation to complete insertion. Insets show plug--socket alignment during motion.

\subsection{Ablation Studies}
\label{sec:ablations}
\textbf{Component effects:} Table~\ref{tab:ablation} denotes the complete \ours\ framework as Full and evaluates the contributions of prior adaptation, value weighting, and successful-action supervision. Uniform W. (uniform candidate weights) replaces Eq.~\eqref{eq:weights} with $w_i=1/N$, retaining the candidate budget and refresh rule. The w/o SFM variant sets online $\beta=0$ while retaining warmup. Frozen ACT disables online behavior absorption. Succ-only Q (success-only critic) restricts value training to successful episodes while keeping the other replay rules. ACT-only removes residual execution, whereas Flow-only directly generates absolute action chunks without an ACT prior. These variants compare policy structures.

\begin{table}[!htbp]
  \centering
  \caption{Ablation results across five tasks.}
  \label{tab:ablation}
  \begingroup
  \footnotesize
  \setlength{\tabcolsep}{1.5pt}
  \renewcommand{\arraystretch}{1.05}
  \setlength{\aboverulesep}{0.35ex}
  \setlength{\belowrulesep}{0.45ex}
  \begin{tabular*}{\columnwidth}{@{\extracolsep{\fill}}lrrrrrrr@{}}
    \toprule
    & \multicolumn{5}{c}{Final success rate (\%)} & \multicolumn{2}{c}{Mean} \\
    \cmidrule(lr){2-6}\cmidrule(l){7-8}
    Variant & Power & USB & Key & Cable & Thread & SR & NAUC \\
    \midrule
    \textbf{Full}
      & \textbf{86} & \textbf{99} & \textbf{99} & \textbf{87}
      & \textbf{68} & \textbf{88} & \textbf{79} \\
    \midrule
    Uniform W.
      & 83 & 95 & 96 & 80 & 56 & 82 & 67 \\
    w/o SFM
      & 80 & 92 & 94 & 73 & 50 & 78 & 67 \\
    Frozen ACT
      & 84 & 97 & 97 & 81 & 60 & 84 & 73 \\
    Succ-only Q
      & 79 & 93 & 94 & 75 & 48 & 78 & 64 \\
    \midrule
    ACT-only
      & 74 & 85 & 88 & 65 & 35 & 69 & 55 \\
    Flow-only
      & 84 & 97 & 97 & 84 & 62 & 85 & 65 \\
    \bottomrule
  \end{tabular*}
  \par\vspace{2pt}
  \begin{minipage}{\columnwidth}
    \scriptsize
    Results are rounded to integers. Bold indicates the best result in each column.
  \end{minipage}
  \endgroup
\end{table}

\ours\ achieves a mean SR of $88$ and NAUC of $79$. Uniform W. reduces these metrics to $82$ and $67$, while w/o SFM gives $78$ and $67$. These results support the complementary roles of value-guided candidate fitting and supervision from observed successful actions. Frozen ACT yields $84$ SR and $73$ NAUC, supporting the benefit of feeding successful corrections back into the prior. Succ-only Q yields $78$ SR and $64$ NAUC, the largest component-wise NAUC drop.

The effects are most pronounced on threading and cable routing. The w/o SFM variant reduces their SR by $18$ and $14$ points, respectively; Succ-only Q reduces them by $20$ and $12$ points. Frozen ACT costs $8$ points on threading and $6$ on cable routing, compared with $2$ on each rigid insertion task. These larger drops highlight the value of behavior retention and critic guidance when manipulating deformable objects.

ACT-only reaches $69$ mean SR and $55$ NAUC. Flow-only achieves $85$ SR, relatively close to Full, but its NAUC is $65$, fourteen points lower. This indicates that removing the action prior particularly affects performance during learning. Since this comparison also changes the action parameterization, it assesses the overall prior--residual structure rather than a single loss term.

\begin{figure}[t]
  \centering
  \includegraphics[width=\columnwidth]{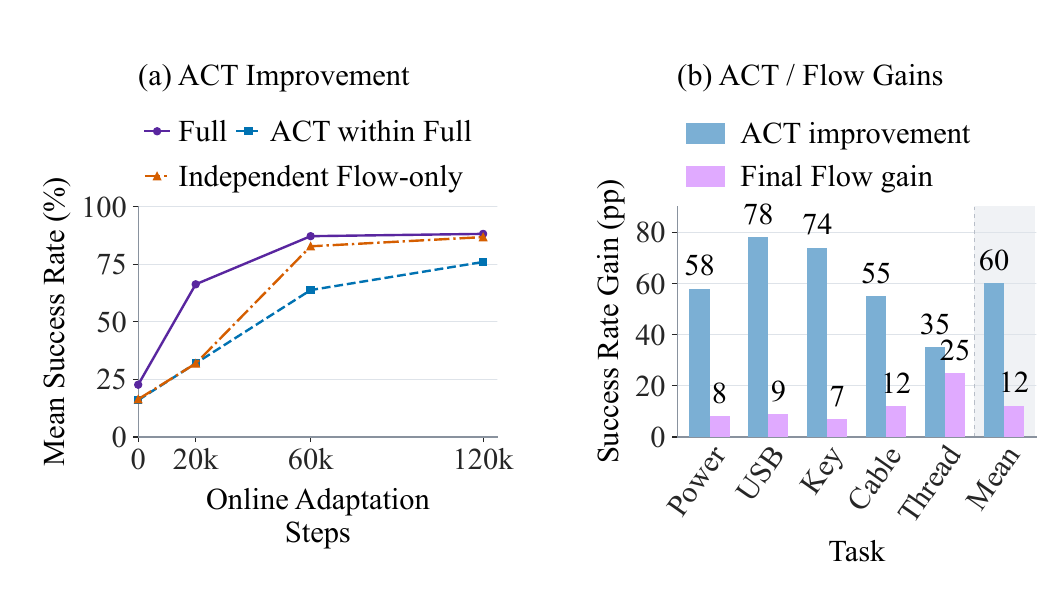}
  \caption{ACT improvement and residual contribution. (a) Five-task mean success at four checkpoints. ACT within Full disables the residual at the same Full checkpoint without retraining; independent Flow-only is separately trained. (b) ACT improvement from $0$ to $120$k and final Full-minus-ACT gain, in percentage points.}
  \label{fig:absorption}
\end{figure}

\textbf{Behavior absorption and residual contribution:} Figure~\ref{fig:absorption} evaluates ACT within Full by disabling the residual at the same combined-policy checkpoint without retraining. This differs from the independently trained ACT-only variant in Table~\ref{tab:ablation}. ACT within Full improves from $16\%$ to $76\%$ mean success, a $60$-point increase, while the combined policy improves from approximately $23\%$ to $88\%$ in this diagnostic. The base-policy improvement demonstrates absorption of successful corrections into ACT. Together with the Frozen ACT ablation, this links the benefit of prior updating to capability retained when residual execution is disabled.

The independently trained Flow-only curve in Fig.~\ref{fig:absorption}(a) further separates early learning from endpoint performance: \ours\ reaches $66\%$ versus $32\%$ at $20$k steps, whereas the respective endpoints are $88\%$ and $87\%$ in this diagnostic. This early lead highlights the action prior's contribution to efficient online adaptation.

Figure~\ref{fig:absorption}(b) shows that the residual remains useful after this absorption. Its final mean gain is approximately $12$ points, ranging from $7$ on key insertion to $25$ on threading. ACT improves most on USB and key insertion ($78$ and $74$ points), whereas threading has the smallest ACT improvement ($35$ points) and the largest remaining residual contribution. A stronger action prior can therefore coexist with a continuing need for correction. Threading illustrates this complementarity: the improved ACT still benefits from residual corrections for precise thread--eye alignment.

\begin{figure}[t]
  \centering
  \includegraphics[width=\columnwidth]{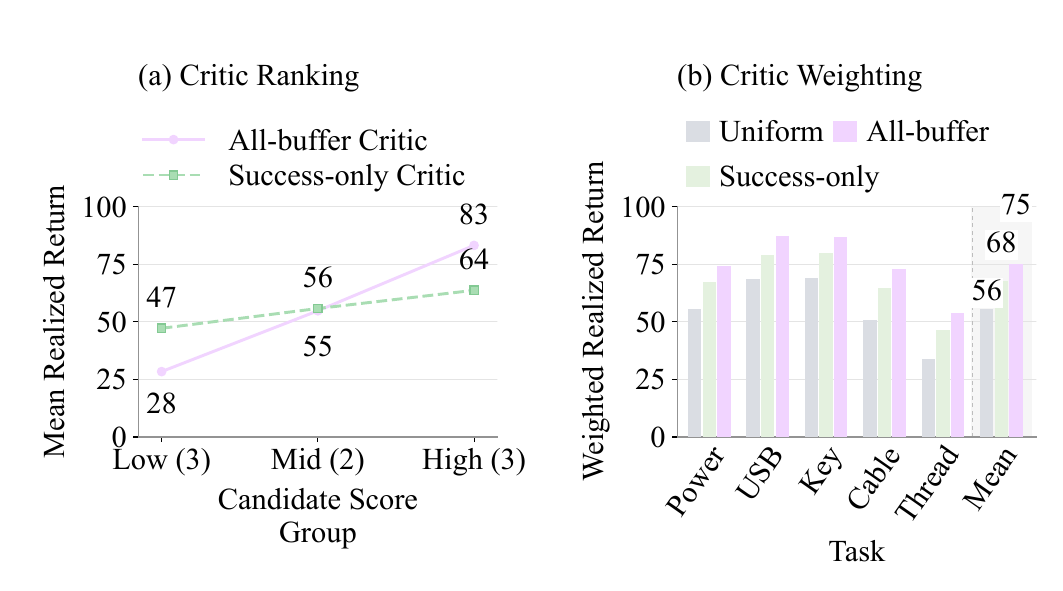}
  \caption{Critic diagnostics on a shared pool of eight candidates. (a) Mean realized returns after each critic ranks candidates into low/middle/high groups of $3/2/3$. (b) Weighted realized returns using uniform, success-only, and all-buffer weights. These quantities measure candidate ranking and weighting quality in return units.}
  \label{fig:critic}
\end{figure}

\textbf{Critic ranking and weighting:} Figure~\ref{fig:critic} compares two critics on the same eight candidates from a frozen ACT--Flow policy. Each ranks candidates into low/middle/high groups of $3/2/3$; group membership may differ. Realized returns use matched initial states and a common continuation policy. Candidate returns are measured in separate evaluations and excluded from training replay buffers.

The all-buffer critic yields low/middle/high mean returns of $28/55/83$, versus $47/56/64$ for the success-only critic. Their high-minus-low gaps of $55$ and $17$ return units show clearer discrimination by the all-buffer critic.

Applying each weighting to the same realized returns gives $\sum_iw_iG_i$, where $G_i$ is candidate $i$'s return. Figure~\ref{fig:critic}(b) reports means of $56$, $68$, and $75$ for uniform, success-only, and all-buffer weighting. All-buffer weighting leads on every task and favors higher-return corrections.

\section{Conclusion}
We presented \ours\ for vision-based insertion. Successful executions improve ACT, while residual relabeling maintains supervision under the evolving prior. Critic-weighted candidate fitting and successful-action FM jointly learn corrections. Five real-world tasks demonstrate efficient early adaptation and behavior absorption, with residuals retaining additional benefits. The resulting policy both learns corrective behavior and incorporates successful experience into its action prior.

The framework requires tasks that admit subtask decomposition with feasible anchor demonstrations and correction warmup. Future work will learn task decompositions and initialize corrections with less reliance on predefined anchors.

\bibliographystyle{IEEEtran}
\bibliography{references}
\end{document}